\documentclass[10pt,twocolumn,letterpaper]{article}

\usepackage{cvpr}
\usepackage{times}
\usepackage{epsfig}
\usepackage{graphicx}
\usepackage{amsmath}
\usepackage{amssymb}

\usepackage{enumitem,kantlipsum}
\usepackage{times}
\usepackage{epsfig}
\usepackage{graphicx}
\usepackage{amsmath}
\usepackage{amssymb}
\usepackage{kotex}
\usepackage{color}
\usepackage{graphicx}
\usepackage{booktabs}
\usepackage{adjustbox}
\usepackage{kotex}
\usepackage{bbold}
\usepackage{pifont}

\usepackage{microtype}
\usepackage{graphicx}
\usepackage{subfigure}
\usepackage{booktabs} 
\usepackage{adjustbox}
\usepackage{kotex}
\usepackage{amsmath,amssymb}
\usepackage{multirow,array}
\DeclareMathOperator*{\argmax}{arg\,max}
\DeclareMathOperator*{\argmin}{arg\,min}
\usepackage{algorithm}
\usepackage{algorithmic}
\usepackage{comment}
\renewcommand{\algorithmicrequire}{\textbf{Input:}}
\renewcommand{\algorithmicensure}{\textbf{Output:}}

\usepackage{hyperref}
\hypersetup{breaklinks=true,bookmarks=false}

\cvprfinalcopy 

\def\cvprPaperID{****} 

\ifcvprfinal\fi

\begin{document}

\title{StackNet: Stacking feature maps for Continual learning}

\author{Jangho Kim\thanks{Equal Contribution}\\
Seoul National University\\
Seoul, Korea\\
{\tt\small kjh91@snu.ac.kr}
\and
Jeesoo Kim\footnotemark[1]\\
 Seoul National University\\
 Seoul, Korea\\
{\tt\small kimjiss0305@snu.ac.kr}
\and
Nojun Kwak\thanks{Corresponding Author}\\
 Seoul National University\\
 Seoul, Korea\\
{\tt\small nojunk@snu.ac.kr}
}

\maketitle
\thispagestyle{empty}

\begin{abstract}
Training a neural network for a classification task typically assumes that the data to train are given from the beginning.
However, in the real world, additional data accumulate gradually and the model requires additional training without accessing the old training data.
This usually leads to the \textit{catastrophic forgetting} problem which is inevitable for the traditional training methodology of neural networks.
In this paper, we propose a {continual learning method}
that is able to learn additional tasks while retaining the performance of previously learned tasks {by stacking parameters}.
Composed of two complementary {components}, {the index module} and the StackNet, our {method} estimates the index of the corresponding task for an input sample with the index module and utilizes a particular portion of StackNet with {this} index.
The StackNet guarantees no degradation in the performance of the previously learned tasks and the index module shows high confidence in finding the origin of an input sample.
{Compared to the previous work of PackNet, our method is competitive and highly intuitive.}
\end{abstract}

\section{Introduction}
The main difference between the human brain and the machine learning methodology is the ability to evolve.
Using neurophysiological processing principles, human brains can achieve and organize knowledges throughout their lifespan.
Having the neuroplasticity, human brains can transfer an activating region of a given function to a different location or control the creation and destruction of synapses according to its experiences.
Usually, artificial neural network (ANN) models consist of a finite number of filters. 
Also, parameters and operations in the ANN do not possess the ability corresponding to the memory system of human brains.
This structural limit leads to the problem called \textit{catastrophic forgetting}, i.e., the newly coming information diverts the model from previously learned knowledge.
The field of researches trying to solve this problem is referred to as \textit{continual learning}.

Many researches based on the regularization method have been proposed to solve this problem.
\cite{jung2017less} and \cite{li2017learning} proposed methods regularizing the output and the feature of the newly trained model respectively.

To make a single network deal with a large number of datasets, 
instead of using all parameters of a single network for each task, PackNet suggested a model which allocates the datasets to specific weights of filters \cite{mallya2017packnet}.
As only the specialized weights for a particular task are involved in its classification process, PackNet shows a remarkable result in multiple tasks.
However, the information about from which task the input comes and which group of weights should be used must be given in advance. 
Typically, images do not contain such prior knowledges and this makes the PackNet hard to apply in real world situation.


\begin{figure*}[t]
  \centering
  \includegraphics[width = 0.6\linewidth]{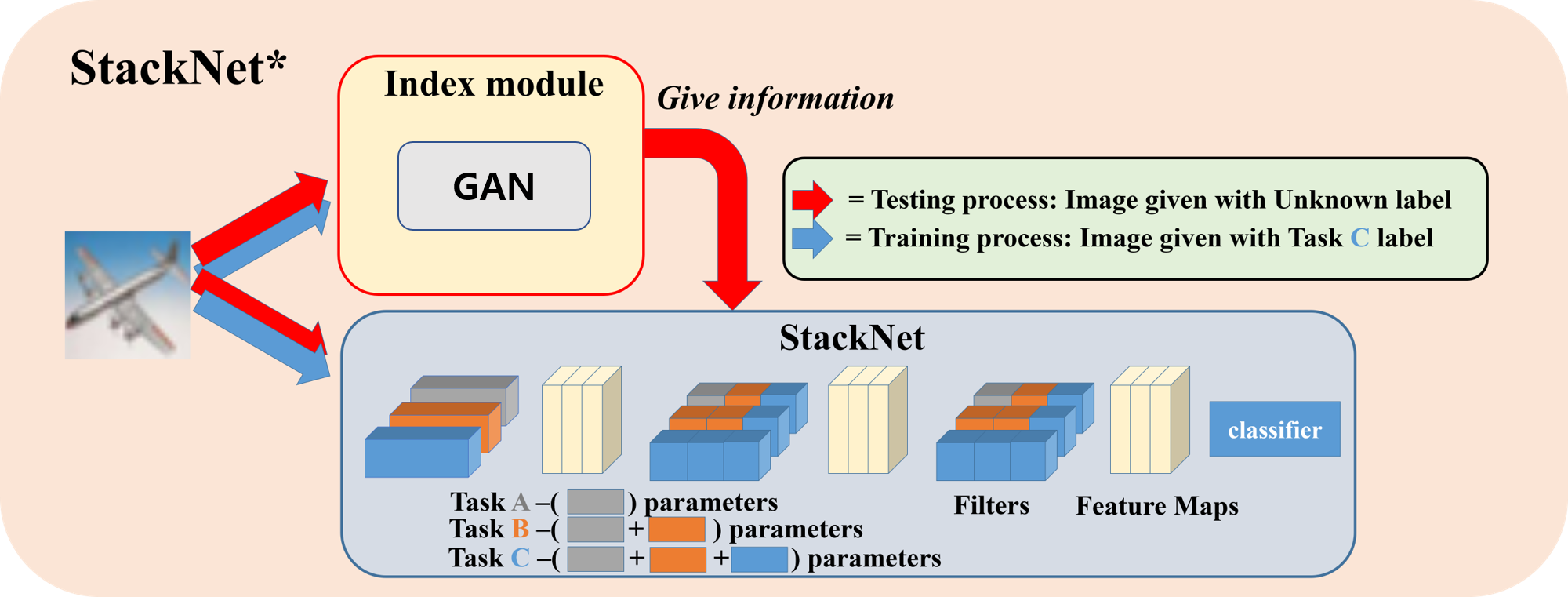}\\
  \caption{The overall architecture of our method called a {StackNet* (Index module + StackNet)}. Blue arrows depicts the training process for the new task (Task C) of our method, where Task A and Task B {have been} already trained. {Index module} is trained each time a dataset for a new task is provided
  and then the {StackNet} is trained {in a standard way such as} using the cross-entropy loss. We allow only the trainable filters (Blue parameters) to be updated. Red arrows shows the inference procedure of our method. {Index module} finds out the task index $\mathcal{J}$ (Task C) from which the input came and let the StackNet to use the filters specialized for task $\mathcal{J}$ (Blue $+$ Orange $+$ Grey). Class labels are also switched to those of task $\mathcal{J}$.}
    \label{ovrall}
\end{figure*}

\begin{figure}[t]
  \centering
  \includegraphics[width = 0.7\linewidth]{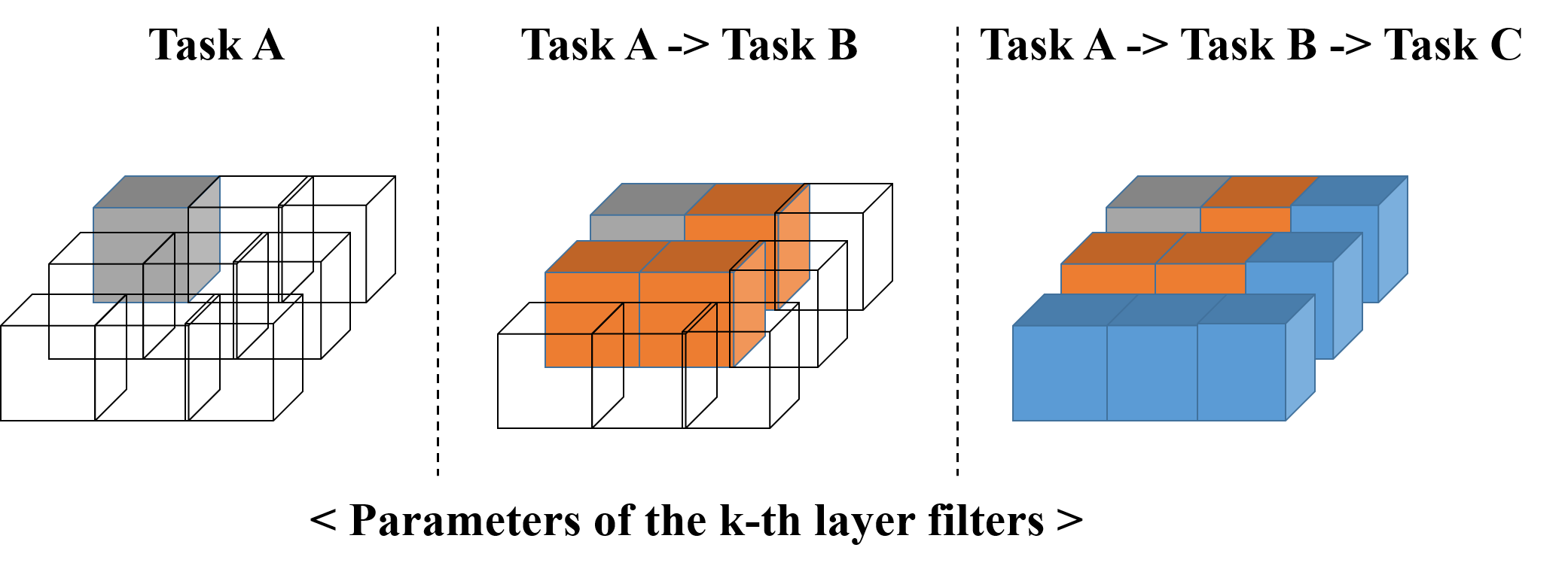}\\
  \caption{{Under the fixed capacity, a certain amount of parameters (Grey) are used in training Task A. After training Task A, additional parameters (Orange) are stacked and trained for Task B. In the same way, The remaining parameters (blue) are trained for Task C.}}
    \label{step}
\end{figure}

In this paper, we propose a network which efficiently {uses the capacity of the network for continual learning} without degrading the performance of {previous tasks}. We have built our {method} using two {complementary components} as shown in Figure \ref{ovrall}.
The \textit{StackNet} keeps the knowledge from several independent tasks.
{After training the StackNet from the previous task with a certain amount of parameters, additional parameters in the convolutional modules are {stacked} and trained with the next task. Note that the given capacity of the StackNet is fixed. StackNet stacks the parameters under the given capacity as depicted in Figure \ref{step}.}
To infer the newly coming data, the model utilizes the newly trained filters along with the previously learned filters, whereas the data from the old task is inferred only using the previously learned filters.
In order to determine which combination of filters to use, we adopted a {index module} which can distinguish the origin of a given input sample. It can recall the information not only which group of filters to use but also which group of class labels to use. This endows our method label-expandability. We suggest the method for {the index module}, which is {generative adversarial networks (GAN) \cite{goodfellow2014generative}} and report the properties and performance of this method.
Our {index module} can be combined with any other works such as PackNet or {Learning without forgetting (LwF) \cite{li2017learning}.}
The combined {index module} and {StackNet} can prevent the \textit{catastrophic forgetting} {using different parameters for different tasks under the constraint network capacity}.
The contributions of this paper can be summarized as the followings:

\begin{itemize}
\item Filters in the convolutional layers are allocated to each task and parameters from the previously learned tasks are shared among post tasks and used for the initialization of post tasks parameters, which makes the overall model compact and efficient.
\item {After using constraint parameters,} attaching parameters whenever a new dataset is trained allows the model to be expandable {as far as the physical constraints, such as memory and processing time constraints, allow.} 
\item To identify which portion of convolutional filters should be used to classify a data, {the index module} is proposed with several methodologies.
\item The overall method is applicable to multi-task continual learning whose number of class labels increases over tasks.

\end{itemize}

\section{Related work}
During the past decade, a lot of progress has been made in numerous fields of machine learning.
Image classification problem {has} induced various useful architectures~\cite{szegedy2015going,he2016deep,simonyan2014very} applicable to any other tasks. {Many} {researchers} {have been} struggling to enhance the performance in object detection or image segmentation problems leaving many masterpieces behind \cite{liu2016ssd,redmon2016you,dai2016r,long2015fully}.
For performance evaluation of the models above, training and test data are usually separated within a single task. However, what we call intelligence must be able to hold the previously-learned knowledge even after learning new
knowledge. In the real situation, the model has to remember
the learned task without seeing the old task data again.

\cite{li2017learning} proposed a {method} trying to retrain {the model by using the outcomes of the previously learned model as anchoring targets.}
When training {a} new task, a considerably small learning rate is used to prevent the model from losing the learned knowledge.
\cite{jung2017less} proposed a method reusing the trained weights of the learned network.
Training a new target network, the feature representation at the last layer is enforced to preserve the response of the old network using $L_2$ regularization.
\cite{kirkpatrick2017overcoming} suggested a penalty on the quadratic distance between the old parameters and the new ones.
These approaches using regularization basically focus on {alleviating} the deformation of the original network.
Therefore, the more the new knowledge flows in{to} the model, the more the performance on the original task degrades.
Moreover, the performance on the new task is not guaranteed to be in its best state.

Dynamic architectures to overcome the \textit{catastrophic forgetting} have been suggested in many researches.
\cite{rusu2016progressive} proposed a progressive network which grows every time a new task is given.
Features in the previous columns are concatenated to the layers of the new column.
progressive network and stacknet seem to be similar such that they can increase their capacity and can learn more continual tasks. However, there are many different things. In the progressive network, each feature is summarized by an additional adapter and passed to the next task network. Therefore, it is not a typical form of CNN and they need to train the additional adapter and increase their capacity with this adapter related specific task. However, the stacknet utilizes the fixed capacity which is given. It uses the task index for splitting the given filters or weights. When the stacknet uses up the given capacity and comes to train a new task, it can then increase the model size beyond the given capacity. 
\cite{mallya2017packnet} proposed a model using the weight pruning method.
Leaving the remaining weights fixed after pruning, the model can solely focus on the task on interest.
Despite of its efficiency and performance, this method, referred as PackNet, requires the prior knowledge of a given input.
Also, once the model is occupied by several tasks, no more task is allowed.
This restricts the model {from being expandable}.
{Unlike the PackNet, our StackNet simply stack parameters in the convolutional filters instead of a cumbersome pruning procedure while allowing the model to expand further.}



{In our method}, instead of saving all the memory in the {main network} and trying to access it directly, the {index module} finds the index of the memory scattered over the {main network} where knowledges are well organized inside.

\section{Method}



We propose a {method that efficiently stacks parameters for continual learning that satisfies three properties as stated below.}

\noindent \textbf{Property 1.}  The \textit{index module} offers the index about {which} part of the \textit{StackNet} should be used.

\noindent \textbf{Property 2.} The {StackNet} learns the knowledge on the new task under the presence of the previously learned knowledge. 

\noindent \textbf{Property 3.} Index module and StackNet are flexible to the expansion of the network {without the retraining of previous tasks} {and as many new tasks can be learned in one network as physical constraints allow.}


\subsection{StackNet}
We propose an efficient way to allocate the capacity of a network according to the given tasks. 
{StackNet} is a network {that can be separated into several partitions for multi-task and }additional filters can be attached {if} a new task is given.
It can be applied to many typical networks such as VGG and ResNet  \cite{simonyan2014very,he2016deep}. 
The only difference with the original versions of those networks is that it uses different parts of filters for different tasks.
As shown in \cite{li2016pruning}, typical convolutional neural networks (CNN) are capable of maintaining their performance even when the majority of their parameters are pruned.
This implies that filters in a CNN contains a lot of unnecessary or redundant information.
Also, sharing the filters among different tasks inevitably leads to the \textit{catastrophic forgetting} of the network since the parameters trained by an old task should be replaced with those trained by a new task.
For these reasons, we utilize a network divided into several parts and refer it as {StackNet} where each part takes charge of a particular task.

\begin{figure}[t]
  \centering
  \includegraphics[width = 0.7\linewidth]{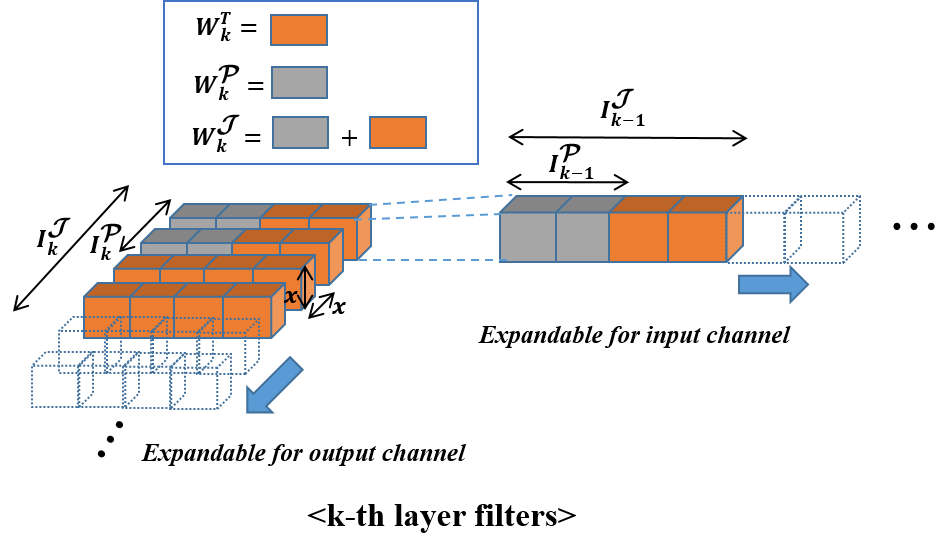}\\
  \caption{Trainable parameters {$W_k^T$ (Orange)} and frozen parameters {$W_k^\mathcal{P}$ (Grey)} in the {StackNet} training process. {In the training process, the parameters (Orange), which are trainable parameters except frozen parameters (Grey), are updated. {$x$ denotes the spatial dimension of filters.}}  }
    \label{cnet}
\end{figure}

While training the {StackNet} with several tasks, each task uses a different part of the {StackNet}.
For the convenience of referring which segments of the {StackNet} are to be activated, we introduce a filter index $I^{\mathcal{J}}$ for the $\mathcal{J}$-th task.
The filter index $I^{\mathcal{J}}$ defines the range of filters to use for the specific task $\mathcal{J}$ in every layers.
For example, if the filter index $I^{1}$ equals to ten, the first task uses filters from the first to the tenth filter.

Depicted in Figure \ref{cnet}, $I_k^\mathcal{J}$ denotes the filter index in the $k$-th layer corresponding to the task $\mathcal{J}$.
This equals to the number of output channels in the {sequent} feature map.
$I^\mathcal{P}_{k}$ denotes the filter index in the $k$-th layer {for the previous task.} 
As the same principle is applied to the precedent ($k-1$)-th layer, the number of input channels {for the filters in the $k$-th layer} will change likewise.
That is, total of $I^\mathcal{P}_{k}$ filters have the {channel} length of $I^\mathcal{P}_{k-1}$ when the previous task is under training.
As the new task is given, the length of the filters become $I_{k-1}^\mathcal{J}$ since the {filters in the prior layer} will deliver a feature map whose length is $I_{k-1}^\mathcal{J}$.
Therefore, both the number of input and output channels increases as the model grows.

Weight parameters of $\mathcal{J}$-th task in filters of the $k$-th layer are referred as $W_k^{\mathcal{J}} \in R^{x \times x \times I^\mathcal{J}_{k-1} \times I^\mathcal{J}_{k}}$, where $x$ means the size of the kernel while $W_k^{\mathcal{P}} \in R^{x \times x \times I^\mathcal{P}_{k-1} \times I^\mathcal{P}_{k}}$ are the weight parameters of the previous task. 
When training the new task, $W_k^{\mathcal{J}}$ is used for the inference but only the parameters {$w \in W_k^{\mathcal{T}}$} ($W_k^{\mathcal{T}} = W_k^{\mathcal{J}} \setminus W_k^{\mathcal{P}}$) are updated while leaving $W_k^{\mathcal{P}}$ fixed.
Note that biases in the filters can also be trained likewise as explained above.
The fully connected layer located just before the linear classifier can be divided as well into the sections of $I^\mathcal{J}$ and $I^\mathcal{P}$.

To enhance the performance, slices of trained parameters $W^{\mathcal{P}}$ from the old task are duplicated to the newly available task parameters $W^{T}$ and Gaussian noise is added to them for the sake of good initialization. 
Then, the network is trained {in a standard way, e.g., using} a cross-entropy loss $\mathcal{L}_c$.
Independent linear classifiers are used for every tasks.
This scheme of {StackNet} can also be applied to structures using shortcut connections such as ResNet. The overall training process is shown in Algorithm \ref{algo-training}.




\begin{algorithm}[t]
\renewcommand{\algorithmicrequire}{\textbf{Input:}}
\renewcommand{\algorithmicensure}{\textbf{Output:}}
\renewcommand{\algorithmicprint}{\textbf{break}}
\caption{\text{{StackNet} Training}}
\label{algo-training}
\begin{algorithmic}[1]
\REQUIRE $W^\mathcal{J},I^\mathcal{P},I^\mathcal{J}$
\ENSURE $W^\mathcal{J*}$
\STATE Freeze the {$W^{\mathcal{P}} \subset W^{\mathcal{J}}$ using the information} $I^\mathcal{P}$
\STATE Initialize the $W^{\mathcal{T}}$ with $W^{\mathcal{P}}$ and a random noise
\STATE $W^\mathcal{J*} \gets \argmin(\mathcal{L}_c$) \\
\COMMENT{Update $W^{{T}}$ using backpropagation}
\end{algorithmic}
\end{algorithm}

\subsection{Index module}
Even if the StackNet is well trained separately with several tasks, {the task index} $\mathcal{J}$ must be given at the time of inference.
Especially when the class labels expands, {this prior information $\mathcal{J}$} is necessary to estimate which group of class labels to use.
In the testing step, existing methods needs the exact origination of the input data.
However in the real world, a given data {for classification} normally does not offer any information about its origination.
Therefore, this prior knowledge should be estimated by an independent network.

To solve this problem, we introduce the {index module} using several different methods, which is able to estimate the task $\mathcal{J}$ of the given input data and inform this to the {StackNet} to specify which filters to use.

\subsubsection{Generative adversarial networks (GAN)}
This method is inspired by the on-line replay method used for continual learning \cite{shin2017continual}.
We train a task-specific generative model $G_\mathcal{J}$ using a GAN to generate pseudo-samples of task $\mathcal{J}$.
The generator is trained using the adversarial loss as follows:
\begin{equation}
\begin{split}
  \min_{G_\mathcal{J}} \max_{D_\mathcal{J}}V(D_\mathcal{J},G_\mathcal{J})= \mathbb{E}_{x\sim P_{data}^\mathcal{J}}[\log(D_\mathcal{J}(x)] + \\ \mathbb{E}_{z\sim p_{z}(z)}[\log(1-D_\mathcal{J}(G_\mathcal{J}(z))].
  \label{equ:gldis}
\end{split}
\end{equation}
Here, $D_\mathcal{J}$ is a discriminator for the task $\mathcal{J}$ and $x$ is a sample from the task $\mathcal{J}$.

After generating samples of all tasks, a task-wise binary classifier $B_\mathcal{J}$ is trained to classify whether the given input is from the task $\mathcal{J}$ or not.
Note that only the generated samples are involved for the training of each binary classifier $B_\mathcal{J}$.
This means that positive samples are from the generator $G_\mathcal{J}$ for the task $\mathcal{J}$ and negative samples are from all the other task generators $G_\mathcal{K}$ ($\forall \mathcal{K} \neq \mathcal{J}$).

In the testing phase, each classifier produces the probability of how likely the given unknown input data is from the task $\mathcal{J}$.
The classifier $B_{i}$ with the highest probability assumes that the input data is from the task $i$.
{index module} can figure out the task $\mathcal{J}$ of input data with maximum probability across the set of classifiers. 
{index module} gives this estimated task index $\mathcal{J}$ to the {StackNet}. 
The index module using GAN can be summarized as below:

\begin{equation}
\begin{split}
&\textrm{Initialization:} \qquad B = \{B_{1}, B_{2}, ... , B_{N}\} \\
&\textrm{Training:} \quad B_{i}(x) = \begin{cases}
1 & \mbox{if $x$ is from $G_{i}$} \\
0 & \mbox{if $x$ is from $G_j$ $\forall j \neq i $}
\end{cases}
\\
&\textrm{Inference:} \qquad \mathcal{J} = \argmax_i  B_{i}(x)
\end{split}
\end{equation}

\section{Experiment}
We evaluate our method on several image classification datasets. 
First, we verify the effectiveness of our method with MNIST \cite{lecun1998gradient}, SVHN \cite{netzer2011reading} and CIFAR-10 \cite{krizhevsky2009learning} which are widely used to evaluate image classification performance. 
Then, we evaluate our method on two subsets of ImageNet \cite{ILSVRC15} which is a real world image dataset.
More details are summarized in Table. \ref{table:dataset}.

We compare our method to various methods such as Learning without forgetting (LwF) and PackNet as well as networks trained for a single target task which shows the performance without conducting continual learning.

\begin{table}
	\centering 
	\caption{
Details of datasets.
}
\begin{adjustbox}{width=1.0\linewidth}
		\begin{tabular}{ | c  || c  c  c| }
			\hline \\[-1.0em]
			Datasets & \#of Train data & \#of Test data & \#of Class\\ \hline \\[-1.0em]
            MNIST & 60,000 & 10,000 & 10\\
            SVHN & 73,275 & 26,032 & 10\\
            CIFAR-10 & 50,000 & 10,000 & 10\\
            ImageNet-A & 64,750 & 2,500 & 50 \\          
            ImageNet-B & 64,497 & 2,500 & 50           
    
    \\ \hline      		
\end{tabular}
\end{adjustbox}
\label{table:dataset}
\end{table}

\subsection{Continual learning {using} basic image classification {datasets}}

We have built {three} experimental scenarios to evaluate our method using basic image classification datasets.
We compare the performance of learning a consecutive {two task} pair among MNIST, SVHN and CIFAR-10 dataset.
After that, we conduct multiple-task continual learning with these datasets.

\subsubsection{Training details}
We use the same learning parameters and network architectures suggested in \cite{jung2017less} used for `Tiny image classification'.
It is composed of three convolutional layers using $5 \times 5$ kernels with the size of {32, 32 and 64} channels respectively, three max pooling layers, one fully connected layer with 200 nodes and the last softmax classifier layer producing 10 outputs. 
ReLU is used as the activation function in all {the} experiments.
Also an SGD optimizer with a mini-batch size of 100 has been used for model optimization.
When training the old task, the weight decay and the momentum were set to 0.004 and 0.9 respectively. 
The learning rate starts from 0.01 and the decay of the learning rate with a factor 0.1 is done at the time of {20,000 iterations}.
After {40,000 iterations}, the training is terminated in all {the} experiments. 
When training {a} new task, LwF, PackNet and {StackNet have the same settings for the old and new task because they need more iterations as the new parameters are to be trained from the scratch. Note that PackNet needs additional pruning and finetuning step. In the new task training of LwF, we start with {a learning  rate of} 0.0002 and 0.001 which are 0.1 $\sim$ 0.02 {times} less than {that of} old task for the `MNIST to SVHN'  and `SVHN to CIFAR-10' respectively, as recommended in \cite{li2017learning}. 
{The channel split ratio which determines $I^\mathcal{J}$ follows the setting of PackNet.}

In Table \ref{table:basic_dataset}, the result of `{StackNet}' shows the performance of the {StackNet} only. That is, it assumes that the {index module} never fails and the task index $\mathcal{J}$ is always given correctly. Likewise, LwF and PackNet need prior knowledge of a given input to know which classifier to use.
As there is no structure like {index module} in the original paper, The actual comparison must be done with `{StackNet}'. On the other hand, {StackNet* shows the performance of the StackNet which uses the task index $\mathcal{J}$ given from the index module.} 

\subsubsection{MNIST $\rightarrow$ SVHN} 
After selecting a subset of SVHN to equalize the number of training data between MNIST and SVHN, images are resized to $28 \times 28$ as in \cite{jung2017less}.
The `single network' in Table \ref{table:basic_dataset} is a network of full capacity trained using a single dataset. 
The column `Old' represents the test accuracy on the previously learn{ed} {task} and `New' is that of the newly trained {task}.
A knowledge distillation loss with hyper-parameters of $T=2$ and $T=1$ \cite{hinton2015distilling} is used for LwF.

LwF can control the performance trade-off between the old and new task by changing the temperature parameter $T$.
PackNet almost maintains the performance of the single network {for the first task} with less than 2\% drop in accuracy.
{StackNet*} outperforms all other methods in the table.
Note that the prior knowledge {to select which task} is given in {all} experiments of {other methods.}
Therefore, {StackNet}, which {shows} the highest in performance, should be compared to other baseline methods.
{Nonetheless, our full model, {StackNet*} also outperforms PackNet, especially on the second task by around 0.7\% {even without using the prior knowledge}.}

\subsubsection{SVHN $\rightarrow$ CIFAR-10}
{In this experiment,  }
LwF and PackNet show similar {trends} as that of `MNIST to SVHN'.
Higher temperature ($T$) induces better performance in the new task but {a bit} of degradation in average performance still occurs.
PackNet experiences almost no performance degradation.
The pruning method {of PackNet} acts like a generalization method and {helps} the model to maintain its capacity.
{The average performance of {StackNet} is slightly lower than that of PackNet by 0.35\%. However}, note that {PackNet} requires additional finetuning after the pruning process {and takes additional time for training.}
{On the other hand,} {StackNet} requires no additional pruning or finetuning precedure and yet retains {the performance.} 

\subsubsection{MNIST $\rightarrow$ SVHN $\rightarrow$ CIFAR-10}



Training three or more tasks is unmanageable for the models using regularization methods.
No matter how we change the temperature ($T$), LwF suffers from a drop in average performance, especially in the third task.
Also, even though the proportion allocated to each task is equivalent between the PackNet and our methods ({StackNet} and StackNet*), our method highly outperforms the PackNet in the third task.
This implies that {StackNet} is more suitable for multi-task sequential learning situation.

\begin{table}
	\centering    
	\caption{
Mean classification results on the basic datasets (5 runs).
}
\label{table:basic_dataset}
\begin{adjustbox}{width=1\linewidth}
		\begin{tabular}{ | c  || c  c  c  c| }
			\hline \\[-1.0em]
			Datasets & Methods & Old(\%) & New(\%)  & Avg. (\%)  \\ \hline \\[-1.0em]
               & single network (MNIST) & 99.49 & --  &-- \\ 
         MNIST& single network (SVHN) & -- & 92.82  &-- \\  \cline{2-5}
     $\downarrow$    & LwF ($T=1$)  & 98.27 & 86.40  &92.34 \\
    SVHN     & LwF ($T=2$)  & 97.33 & 86.97  &92.15\\
         & PackNet  & 99.45 &  91.49 &95.47\\

& {StackNet*}   & 99.43 & 92.20  & 95.82 \\  
         & {StackNet}   & 99.43 & 92.37  & 95.90 \\ \hline 
                & single network (SVHN) & 92.94 & --  &-- \\ 
         SVHN& single network (CIFAR) & -- & 79.69  &-- \\  \cline{2-5}

         $\downarrow$ & LwF ($T=1$)  & 91.76 & 70.57  &81.17 \\
         CIFAR-10 & LwF ($T=2$)  & 90.19 & 71.94  &81.07 \\
                  & PackNet  & 92.84 & 76.78  &84.81\\

         & {StackNet*}   & 90.05 & 76.62  & 83.34 \\  
        & {StackNet}  & 92.21 & 76.70  &84.46 \\ \hline 

\end{tabular}
\end{adjustbox}
\begin{adjustbox}{width=1\linewidth}
		\begin{tabular}{ | c  || c  c  c  c  c| }
			\hline
			Datasets & Methods & MNIST(\%) & SVHN(\%)  & CIFAR(\%) & Avg(\%)   \\ \hline
              & single network (MNIST) & 99.44 & --  &-- &--\\ 
         MNIST       & single network (SVHN) & -- & 92.94  &-- &--\\ 
$\downarrow$  & single network (CIFAR) & -- & --  &79.69 &--\\  \cline{2-6} 
    SVHN & LwF ($T=1$) & 95.06 & 86.80  &68.98&83.61 \\ 
    $\downarrow$ & LwF ($T=2$) & 86.09 & 85.07  &69.63&80.26 \\ 
     CIFAR-10        & PackNet  & 99.38 & 91.93  &66.34&85.88\\

         & {StackNet*}  & 99.41 & 89.36  & 74.93  &   87.90 \\  
          & {StackNet}  & 99.41 & 91.84  &75.02&88.76 \\ \hline 
\end{tabular}
\end{adjustbox}
\end{table}

\subsubsection{Multitasks more than three tasks}

We also conducted an experiment for 5 tasks to verify effectiveness on multitasks more than three tasks. We used 5 datasets, cifar10, svhn, KMNIST \cite{clanuwat2018deep}, FashionMNIST \cite{xiao2017fashion} and MNIST. We used ResNet-32 \cite{he2016deep}for the base network. The performance of single networks are 86,95,98,93 and 99 respectively. In the same order, the performance of the stacknet are 83.93,95,91 and 99. As we can see, the results have a similar tendency compared to the previous experiments. The accuracy degradation is within the range of 0~3\%. Also, StackNet can handle more than 3 tasks.

\subsection{Continual learning using realistic datasets}
ImageNet contains images more realistic than other datasets used in this paper.
Higher resolution and complex backgrounds make the classification even harder.
To save the training time, two subsets of the ImageNet dataset each having 50 randomly chosen classes have been used for evaluation.
They are referred to as ImageNet-A and ImageNet-B respectively in this paper.
To show the adaptability of our method on structures having shortcut connections, ResNet-50 is used for the experiment. {We use the same experimental setting as in \cite{he2016deep} with a batch size of 128.}
We compare our {StackNet} with LwF and PackNet.

Table \ref{table:real-world} shows the results on the ImageNet dataset.
The `Old' accuracy of all methods slightly dropped from those of single networks.
Like in the `SVHN to CIFAR-10' experiment, PackNet shows a better result than the LwF regardless of the value of $T$.
The result of {StackNet} is almost identical to PackNet with a slight increase in the average accuracy.


PackNet actually utilizes more parameters for each task than {StackNet} does.
Since the masks force weights with no influence to be zero and utilize the well trained remaining weights, PackNet can make good use of the entire network.
However in {StackNet}, just adding filters gradually without any other post-processing performs well enough compared to PackNet.
When the model of an initially designed size is fully occupied by several tasks, {StackNet} can just add more filters and train them along with the trained filters while this is not the case of PackNet.
Furthermore, PackNet requires additional memories to store the masks for all filters in it.
{StackNet} needs only one integer per layer for each task.
For these reasons, {StackNet} is far more efficient than the PackNet with no loss in performance.

\begin{table}
	\centering    
	\caption{
Mean classification results for the realistic dataset (2 runs)
}
\label{table:real-world}
\begin{adjustbox}{width=1\linewidth}
		\begin{tabular}{ | c  || c  c  c  c| }
			\hline
			datasets & Methods & Old(\%) & New(\%)  & Avg. (\%)  \\ \hline
               & single network (ImageNet-A) & 83.28 & --  &-- \\ 
         ImageNet-A & single network (ImageNet-B) & -- & 85.28  &-- \\  \cline{2-5}
         
         $\downarrow$ & LwF ($T=1$)  & 82.2 & 86.72  &84.46\\
         ImageNet-B & LwF ($T=2$)  & 80.92 & 86.96  &83.94 \\
         
         & PackNet   & 82.16 & 88.72  &85.44\\

& {StackNet}  & 83.30 & 88.66  &85.98 \\        
\hline

\end{tabular}
\end{adjustbox}

\end{table}

         
         



\subsection{Ablation study on the efficiency of \textbf{StackNet}}
As mentioned above, our StackNet appends {parameters} whenever a new task is to be trained.
In this process, parameters learned from previous tasks are utilized altogether and the parameters which are to be trained newly are initialized using the existing ones.
We conduct experiments to show the effect of these methods.
Experiments are carried out in `MNIST to SVHN' case as a representative.

\subsubsection{Initialization with pretrained parameters}
To analyse the effectiveness of initialization using pretrained old task parameters with random Gaussian noise, we compare the accuracy between a model initializing the new parameters with just a random Gaussian distribution and a model initializing the new task parameters using the pretrained old task parameters with additional random Gaussian noise.
In Table \ref{table:ablation}, the model initialized by our method obviously shows higher accuracy on the SVHN task.
This result implies that a good initialization prevents the model from going through a local minima.

\begin{table}
\begin{center}
\caption{ 
{Performances of the old (MNIST) and new (SVHN) tasks with different initialization methods: 1) initialization with only random Gaussian noise and 2) the pretrained parameter added by random Gaussian noise}
}
\label{table:ablation}    

\begin{adjustbox}{width=0.8\linewidth}
		\begin{tabular}{ | c  || c  c  c| }
			\hline
			\ Initialization & Old(\%) & New(\%) & Avg.(\%)   \\ \hline
  Random & 99.42 & 91.67 &95.55\\
 Pretrained parameters & 99.43 & 92.37 &95.90\\ \hline
  \end{tabular}
		
\end{adjustbox}

\end{center}
\end{table}

\subsubsection{Parameter sharing}
We verify the effect of the parameter sharing between the old task and the new task.
The baseline method is a model where the old parameters are filled with random Gaussian initialization and no further training is done.
The new parameters in the model have to learn knowledges from the new task without the aid of old parameters since they are fixed from the beginning.
On the other hand, the old parameters in our method is trained by the old task and the new parameters can make use of these learnt knowledges to learn the new task.
To solely observe the effect of the parameter sharing, both {methods do} not use our parameter initialization method mentioned in the previous section.
The task indices $I^1$ and $I^2$ are \{16,16,32\} and \{18,18,34\} respectively.
The model using parameter sharing converges far more faster than the model with no parameter sharing which is shown in {Figure} \ref{param}.
Also the final error rate of our method is lower than the other.
Increasing the number of indices elevation enhances our method by 6.38\% and the other model by 7.91\% as shown in Table \ref{table:ablation2}.
This implies that the parameter sharing improves the model and allow it to be compact with fewer numbers of filters.

\begin{figure}[t]
  \centering
  \includegraphics[width = 0.8\linewidth]{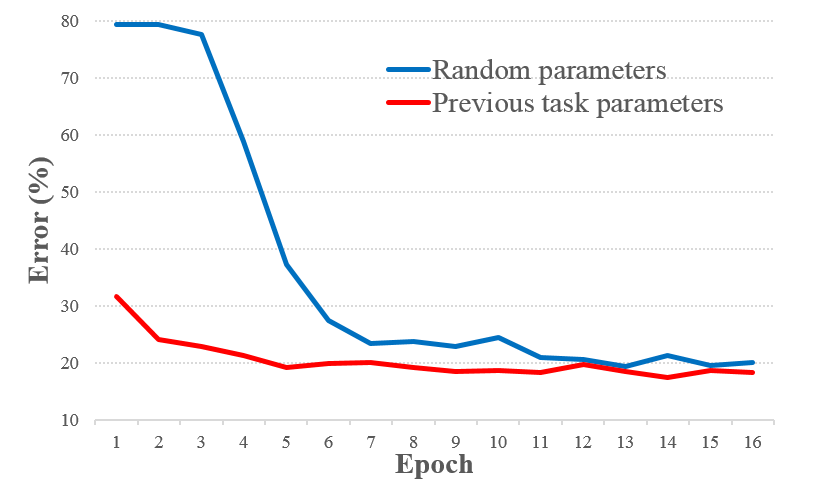}\\
  \caption{Error rate comparison between sharing the parameters which learned previous task and random parameters. }
    \label{param}
\end{figure}

\begin{table}
	\centering   
\caption{ 
The performances on the new task while increasing the filter index of the model using parameter sharing and the model with random fixed parameters.
}
\label{table:ablation2}
\begin{adjustbox}{width=0.8\linewidth}
		\begin{tabular}{ | c  || c  c | }
			\hline
			\#of increasing each index & Ours(\%) & Random(\%)    \\ \hline
2  & 80.60 & 78.10 \\
4 & 85.31 & 83.85 \\
6 & 86.98 & 86.01 \\ \hline
  \end{tabular}
	\end{adjustbox}

\end{table}

\begin{table}
	\centering    
	\caption{
{Experimental results of the index module on the basic datasets}
}
\label{indexModule}
\begin{adjustbox}{width=1\linewidth}
		\begin{tabular}{ | c  || c  c  c  c| }
			\hline \\[-1.0em]
			Datasets & Methods & Old(\%) & New(\%)  & Avg. (\%)  \\ \hline \\[-1.0em]
               MNIST$\rightarrow$   & GAN & 100 & 99.94  &99.97 
           \\  SVHN&  Baseline & 61.65 & 97.17  &79.41     \\  \hline
    SVHN$\rightarrow$  & GAN & 97.00 & 99.90  &98.45\\
       
   CIFAR-10  &  Baseline & 89.39 & 85.86  & 87.63 \\
    \hline     
\end{tabular}
\end{adjustbox}
\begin{adjustbox}{width=1\linewidth}
		\begin{tabular}{ | c  || c  c  c  c  c| }
			\hline \\[-1.0em]
			Datasets & Methods & MNIST(\%) & SVHN(\%)  & CIFAR(\%) & Avg(\%)  \\ \hline \\[-1.0em]
   MNIST$\rightarrow$ SVHN  & GAN & 100 & 96.65  &99.83 & 98.83 \\
          CIFAR-10    & Baseline & 52.88  & 90.89  & 75.7 & 73.16 \\
     \hline 
\end{tabular}
\end{adjustbox}

\end{table}

\subsection{Index module}
To solely examine the influence of index module on our method (StackNet*), we have conducted experiments with the same experimental scenarios in StackNet* for GAN. {As a baseline method for the index module, we adopted a highly naive method that chooses the task index $\mathcal{J}$ according to the confidence of the StackNet.
Inferring each task output from the StackNet, the output with the highest probability decides the task index.}
(e.g, Task index $\mathcal{J} = \argmax_i(P_i)$ where $ P_i$ is the highest probability of $i$-th classifier)

In continual learning on MNIST to SVHN and SVHN to CIFAR-10 datasets, an index module approximates the results to perfection (See Table \ref{indexModule}). {The experiment of three datasets also shows the same trend as the experiments above.
The baseline definitely shows a poor performance also in all cases, which underlines the necessity of the index module.} The generated samples with GAN are depicted in Figure \ref{GAN_mnist},\ref{GAN_svhn} and \ref{GAN_cifar}.

\subsubsection{Limitation}
Generative models usually have difficulties in generating realistic images with high resolution and there are few researches experimented on ImageNet data.
This incompetence makes the index module hard to be applied to the ImageNet data.
{Also, when it comes to dealing with datasets of similar distribution, the index module may not be working as good as it is reported in this paper.}

\begin{figure}[h]
\centering     
\subfigure[Generated samples with the generator of index module]{\label{fig1:a}\includegraphics[width=0.43\linewidth]{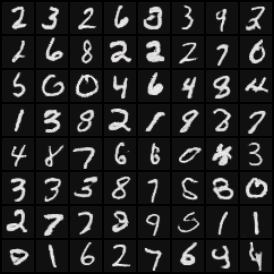}}
\subfigure[Real samples]{\label{fig:b}\includegraphics[width=0.43\linewidth]{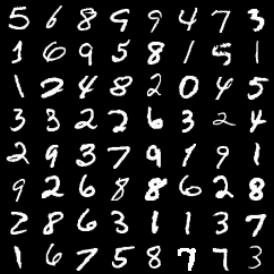}}
\caption{MNIST dataset}
\label{GAN_mnist}
\end{figure}

\begin{figure}[h]
\centering     
\subfigure[Generated samples with the generator of index module]{\label{fig2a}\includegraphics[width=0.43\linewidth]{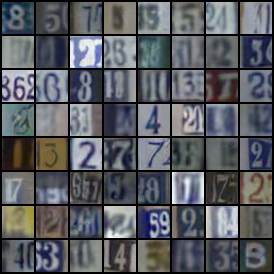}}
\subfigure[Real samples]{\label{fig:b}\includegraphics[width=0.43\linewidth]{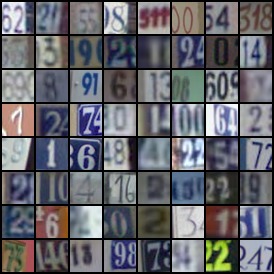}}
\caption{SVHN dataset}
\label{GAN_svhn}
\end{figure}

\begin{figure}[h]
\centering     
\subfigure[Generated samples with the generator of index module]{\label{fig3:a}\includegraphics[width=0.43\linewidth]{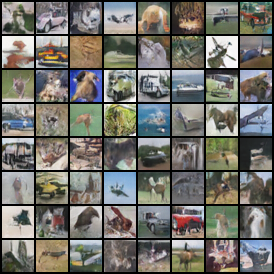}}
\subfigure[Real samples]{\label{fig:b}\includegraphics[width=0.43\linewidth]{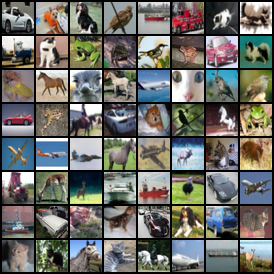}}
\caption{CIFAR-10 dataset}
\label{GAN_cifar}
\end{figure}

\section{Conclusion}
In this paper, we have proposed a novel framework which is able to divide its capacity into several parts and utilize them according to the given input.
Composed of two networks, the index module is responsible for memorizing ``where the data came from'' while complicated knowledge such as ``what the data is'' is engraved to the StackNet.
To the best of our knowledge, our work, index module, is the first attempt to estimate the origin of data which used to be assumed as given in the previous works.
As well as overcoming the catastrophic forgetting, StackNet allows extra class labels when training a new dataset and also has expandability to the network architecture.

\subsubsection*{Acknowledgments}
{This work was supported by Next-Generation Information Computing Development Program through the NRF of Korea (2017M3C4A7077582).
}

{\small
\bibliographystyle{ieee_fullname}
\bibliography{egbib}
}

\end{document}